# Knowledge Graph semantic enhancement of input data for improving AI

Shreyansh Bhatt, Amazon[1]
Amit Sheth, AI Institute, University of South Carolina
Valerie Shalin, Wright State University
Jinjin Zhao, Amazon

**Abstract**
Intelligent systems designed using machine learning algorithms require a large number of labeled data. Background knowledge provides complementary, real world factual information that can augment the limited labeled data to train a machine learning algorithm. The term Knowledge Graph (KG) is in vogue as for many practical applications, it is convenient and useful to organize this background knowledge in the form of a graph. Recent academic research and implemented industrial intelligent systems have shown promising performance for machine learning algorithms that combine training data with a knowledge graph. In this article, we discuss the use of relevant KGs to enhance input data for two applications that use machine learning -- recommendation and community detection. The KG improves both accuracy and explainability.

Machine learning algorithms trained with a large labeled data have shown promising performance in solving problems from various domains [1]. One of the most challenging aspects associated with using such algorithms is the availability of training data. On the other hand, symbolic knowledge representation has been a key area of Artificial Intelligence research since the mid 1970s, yielding a number of vetted background knowledge bases. The AI community started to use the term "Ontology" in the 1980's to refer to such background knowledge [2]. A patent filed in 2000 described the use of background knowledge to power commercial faceted and semantic search, semantic browsing, semantic personalization and semantic advertisement [3]. Subsequently, background knowledge has played a key role in various tasks ranging from search and classification to personalized recommendations. In this decade, several researchers have explored the role of background knowledge to enhance natural language processing and machine learning [17].

In this paper we first describe the history of KGs and their application in research and industry, and introduce the problem of augmenting training data with the contents of a KG. We then review the most common approaches to augmenting data with knowledge, contrasting simple, explicit association of graph content with input data and approaches that depend on deep learning to combine separate KG and input content. We list the challenges associated with these and provide an overview of an alternative joint optimization based approach for KG enhanced machine learning. We report four case studies in different domains that used this approach.

**History**
Google introduced its "Google Knowledge Graph" in 2012, acknowledging its central role in their entity search[1]. Although a number of knowledge representation alternatives have been used, the graph has been one of the most popular formats to represent domain knowledge[2]. Apart from Google Search, a number of commercial products such as Apple Siri,and Amazon Alexa are powered by a KG.
A KG is a collection of facts where entities (nodes) are connected with typed relationships. The scope of the knowledge captured by a KG can vary with broad based coverage that may involve many generic domains (e.g., DBpedia and Yago), a specific domain (e.g., Bio2RDF and UMLS for aspects of biomedical or medical domains), an industry or an enterprise [18]. To extract a KG for a domain of interest [4] domain specific KG creation approaches start from one or more entities (concepts) in the KG. The graph of interest is then created by traversing from those initial entities. However, the method does not gener as even a 2-3 hop traversal may end up including more than 50% of the source content [4]. To control traversal for generic sources, traversing is restricted through relevant relationships (edges) that can be identified by computing a specificity score of relationship to the domain. Research in creating domain specific sub knowledge graphs has shown promising applications [4].

---
[1] Work done prior to joining Amazon

KGs have potential applications in augmenting training data for machine learning algorithms. Training data augmented by a KG has been shown to enhance performance of applications that have limited training data such as sentiment analysis, named entity recognition, recommendation, question answering, and object detection [5, 6, 7, 16].However, the training data may not be available in the same form as the KG,, hindering a facile augmentation of training data.

## Simple data elaboration using a KG

KGs provide auxiliary factual information about the entities that are present in the training data. A simple approach to augmenting training data is to enhance the training data with auxiliary information extracted from the KG [18]. Consider a sentiment classification task on tweets. If a tweet mentions the president of the the USA, a KG such as DBpedia has the information that Donald Trump is the current president of the USA. The input data augmentation approach enhances the tweet representation with concepts associated with the president of the USA in DBpedia. One of the early approaches for sentiment analysis used this strategy and reported an F1-score improvement with tweets. Specifically, for each extracted entity (e.g. iPhone) from tweets, this approach adds its semantic annotation (e.g. "Apple product") as an additional feature, and measures the correlation of the added concept with negative/positive sentiment [5]. Treating the concepts obtained from the KG as one of the tweet features results in a 6.5% increase in the F1-score for sentiment classification.

KGs provide rich information that not only includes a type associated with the concept but also other related concepts. Training data augmentation with KG content requires relative weighting of the following:
1. Concepts present in the training data
2. KG concepts that map to the concepts in the training data, e.g. Tweet about Donald Trump maps to the `dbpedia:Donald Trump` in the DBpedia KG.
3. KG concepts that are associated with different types of relationships with the concept in the training data. E.g., `dbpedia:President` of the United States is associated with Donald Trump in DBpedia with `dbpedia:type` relationship.

## Augmented deep learning with KGs

Separate training data, knowledge concepts, and related concepts from KGs can be input to the neural network, which finds the appropriate importance of these different modalities. Artificial neuron patterns or neural network architectures that compound training data, KG concepts, and related concepts can differ depending on the nature of the training data (text or user-item interaction and time series or static) . Most of the approaches use the first input layer of the deep neural network architecture as the layer that augments training data with the KG. The remaining layers are application and task specific with loss computed at the last layer of the deep neural network. The end-to-end training of such a network results in learning the relative weighting between the training data and different concepts from the KG to solve the application or task such as sentiment analysis, machine reading, recommendation etc. A key advantage of using the neural network to augment the training data with the KG is that a neural network can handle the nonlinearity involved in merging the training data and the KG. A set of artificial neurons, so-called neural network layer, considers the input from different modalities of data. Each neuron on this layer combines these different inputs and applies a (non-linear) activation function. Hence, deep network with multiple layers can learn the appropriate non-linear combination of the training data and the KG. This approach has shown promising results for various applications such as sentiment analysis, recommendation, machine reading and collaborative filtering.

**Augmented deep learning for sentiment analysis:** Kumar et al. proposed an approach to sentiment analysis that augments an input text word with concepts from WordNet KG [6]. The proposed model takes a sentence as input to a BiLSTM that computes a hidden representation of words in the sentence. The hidden representation is then passed through an attention layer along with the KG concepts related to the word. The attention layer then computes a weighted vector of the hidden representation of the input word and KG concepts. Weighted vectors of a sentence are passed through another attention layer, the output of which predicts sentiment. The whole network is trained to predict the correct sentiment of a given sentence in the training data.

**Personalized news recommendation using KG:** Wang et al. reported that a KG plays a key role in personalized news recommendation [7]. During the inference, the model predicts a click through rate for a given user and a given news story. The model generates user representations from their prior history of news click. The user representation is then concatenated with a given news story's representation to generate a *user* context vector. The resulting user context vector predicts the click through rate. For training, authors represented each news story with the entities found in the news story and context (neighborhood of the entity in the KG) of each entity. A multi channel representation is used for each word of a news story that corresponds to an entity in the KG. For example, one channel corresponds to the KG's representation for the entity and another channel corresponds to a representation of related entities. A convolutional neural network is then applied on such representation that combines word level information with the KG's information. Such a representation of each news story is then combined using an attention network to generate a user representation which in turn is combined with the candidate news story's representation to predict the click through rate.

**Machine translation using KG:** One of the challenges in using concepts from a KG to augment text data is a relative weighting of the actual word in the sentence and the word's representation from the KG. Yang et al. reported that while using background knowledge for machine reading, it is crucial to have both the relative weighing of word in the text and its KG based representation, as in some cases the text context properly overrides the context-independent background knowledge available in KG [8]. They use a sentinel vector that combines the word and its related concepts in the KG. Their model uses a BiLSTM where the hidden representation from each BiLSTM unit is combined with related concepts from the KG corresponding to that word using the sentinel vector. The resulting vectored representation is used as the BiLSTM cell's hidden representation. Training this network results in learning the weights for the sentinel vector.

**KG for recommendation:** Zhang et al. showed that a KG can be used to solve the data sparsity issue arising from collaborative filtering for item recommendation [9]. They use an item's representation computed from the KG in user-item feedback matrix. This item's representation is computed by concatenating the visual and textual item's representation available in a KG. They showed that such a combined item's representation in the user-item matrix can improve collaborative filtering for recommendation. Other work on augmenting training data with a KG in recommendation such as the Personalized Entity Recommendation [10] and Factorization Machine with Group lasso [11] treat KG as a heterogeneous information network, and extract meta-path/meta-graph based latent features to represent the connectivity between users and items along different types of relation paths/graphs.

## Challenges for KG augmented machine learning

As noted above, the data format of the input data and KG are often different from each other and require different processing algorithms and architectures. For example, most NLP tasks have sentences as the input data while background knowledge bases are available in graph form. To augment the input data with a KG requires either converting the knowledge into the format of input data or representing input data in the KG.

Differences in processing algorithm and architectures mean that augmenting the input data with a KG and converting into a common format may not lead to the best representation.KGs represent different kinds of information about a concept indicated by different types of relationships. For example, in DBpedia the concept dbr:ohio is connected with the concept dbr:USA by a hierarchical relationship dbo:country while dbr:Ohio is connected to columbus with dbo:Capital relationship. The algorithms exploiting the KG must be cognizant of the different types of relationships. Moreover, these algorithms should find weights appropriate for the different relationships for the given task instead of using a generic relationship weighting of learned for a KG completion task.

Moreover, as the knowledge is fused with the training data before applying a machine learning algorithm with non linearity, the KG may not facilitate explainability.

To address these challenges, recent approaches propose specialized algorithms or neural network architectures for the input data and for KG. We review these approaches as Iterative optimization for enhanced machine learning using a KG.

## Iterative optimization for knowledge enhanced machine learning

In order to augment input data effectively with a KG and to preserve the explainability potential of the KG, recent approaches iteratively optimize the task specific objective for the input data and for the KG representation of the data [12, 13]. As shown in Figure 1, these approaches start with an initial KG representation. The input data is augmented with the initial KG representation. Optimizing the application-specific objective on such input data provides application-specific results. The knowledge graph is updated based on the results and the application-specific objective optimization is then run for a KG, which leads to an improved (application-specific) representation of the KG. The whole process is repeated until convergence or a predefined number of epochs. The initial generic KG representation augments the input data and iteratively updates the KG representation. Next, we review four approaches designed based on this concept.

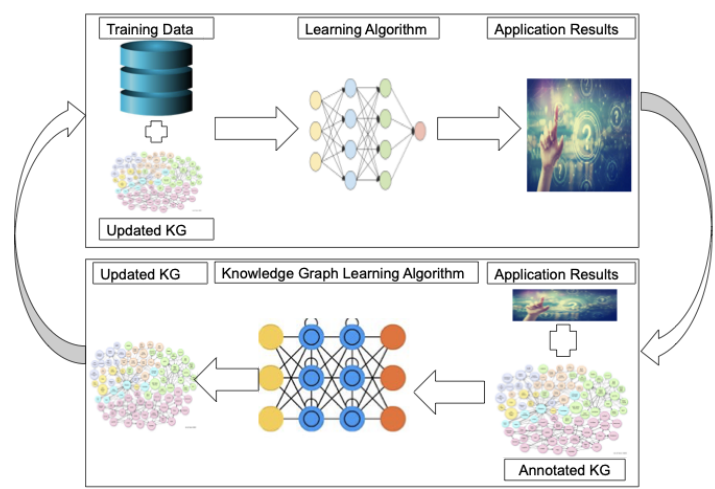

**Figure 1**: Iterative optimization for knowledge enhanced machine learning. Training data is linked/augmented with the KG. The learning algorithm is applied on the augmented data to find the results. The results then drive the annotated KG based learning to identify the updated KG. The updated KG is then used in the training data augmentation.

**KG enhanced recommendation:** Wang et. al. proposed a multi-task feature learning approach for KG enhanced recommendation [12]. For a given user-item interaction matrix, the item's representation is initially computed from the KG. A collaborative filtering technique is then applied on such a user-item interaction matrix. As a second step, the user-item interaction is represented in the KG with a concept specific to the item in KG and the other related item based on user-item matrix. The KG based recommendation is then solved for predicting item similarity. Iterative optimization of these two objectives leads to learning an application specific KG representation that can be used for explanation and also to enhance performance for collaborative filtering. Wang et al. proposed to use "cross and compress" units for combining a KG's representation to item and user-item representation to KG.

**KG enhanced community detection and characterization:** KGs can also help improve our understanding of networked structured data (graph structured data) such as social networks. An attributed graph consists of nodes, attributes associated with nodes, and relationships (links) that connect nodes. For example, in a social network, users are nodes, location or user posts are attributes, and users are connected with friendship relationships. A group of users form a community when the number of relationships within a group exceeds the number of relationships across a group. It is often hard to divide a graph in communities. Node attributes can help explain certain communities. However, communities are often formed because a group of users share a generic concept. For example, a group of users may be friends with each other as they live in the same county. Readily apparent user attributes such as a city name may not inform the county characterization for the group. Crowdsourced KGs such as DBPedia have the information that connects cities to a county.

Bhatt et al. proposed KG enhanced community detection [13]. They used iterative optimization over an attributed social network graph and a hierarchical KG to detect and characterize communities. The hierarchical KG can represent real world communities or clusters. For example, a hierarchical KG for the geo-location domain represents the united states of america as a root concept with California and Ohio as subsuming concepts. Attributes on the nodes of the social network graph are mapped to the KG. Hence, each node is represented in the KG. Initially, each node is considered to be in its own community with a relationship weight computed according to the distance between nodes in the hierarchical KG. In each iteration, first the community detection objective is applied on the graph. Depending on the communities identified in the graph, the hierarchical KG is broken into multiple hierarchical KG representing each community. Hence, the communities identified in the graph inform KG representation of nodes and the input graph is modified based on the distance of nodes in the modified hierarchical KGs.

**Unknown relevant domain:** The initial mapping of the input data to the KG may represent multiple, or unknown domains. However, the objective optimization for the input training data may only depend on a certain domain or an intersection of multiple domains. Hence, if we augment the KG with the training data prior to optimizing the application specific objective we may miss the appropriate background knowledge domain.

Social media-based wisdom of crowd analysis is one of the application domains where the objective optimization on the input data depends on the appropriate domain in the KG. Recent research shows that diverse crowds bring diverse perspectives in decision making [14]. Such a decision results in a more accurate forecast than a decision made by a randomly selected or homogeneous crowd. As users share their opinion on social media, we can use social media data to infer diverse crowds.

Diverse crowd selection can be solved as subset selection, maximizing diversity within the subset. As we want to measure diversity in perspectives in the given domain of interest, we can use data on individual users' attribute and augment it with the KG. However, it is hard to identify the domain of interest from the KG in the context of selecting a diverse crowd. Hence, we can find the diverse crowd by starting with a generic KG based user attribute augmentation and then find the appropriate domain of interest for the given set of diverse crowd. For example, a crowd may be diverse in the domain of politics while it may not be diverse in the domain of sports. This results in the new domain of interest for diverse crowd selection and helps identifying the appropriate diverse crowd.

**Generative applications:** Domain specific short text generation suffers from limited training data. Short and diverse text generation can benefit from domain knowledge. Recent research shows that domain knowledge captured in the form of word2vec vectors improves text generation quality [15]. Here, the specific training data can be limited while the data to generate word2vec vectors can consist of signals related to generic text generation such as grammar and sentence structure. The use of domain specific KGs can further improve diverse text generation quality as it captures words and rules associated with the domain. However, it is challenging to identify the appropriate domain in the KG that helps the particular text generation. Iterative optimization can help such diverse text generation.

## Conclusions

KGs play a key role in machine learning. Crowdsourced KGs can complement the available training data for machine learning algorithms and improve performance for a number of applications. Iterative optimization can further improve accuracy and also help explain the data in the context of the application. This approach is particularly useful when the entities present in the input data are associated with a concept in knowledge graph that is present in the multiple domains. An iterative approach can identify the appropriate domain in the context of the application.

Acknowledgements: We thank Manas Gaur and Ruwan Wickramarachchi for their review and suggestions. Drs. Sheth and Shalin's work was funded in part by NSF award #1513721 titled "TWC SBE: Medium: Context-Aware Harassment Detection on Social Media." Any opinions, findings, and conclusions or recommendations expressed in this material are those of the authors and do not necessarily reflect the views of the National Science Foundation.

## References

[1] Halevy, Alon et al. The unreasonable effectiveness of data. IEEE Intelligent Systems. 2009;24
[2] Gruber, Thomas R. "Toward principles for the design of ontologies used for knowledge sharing?." *International journal of human-computer studies* 43.5-6 (1995): 907-928.
[3] Sheth, Amit at al. "System and method for creating a semantic web and its applications in browsing, searching, profiling, personalization and advertising." U.S. Patent No. 6,311,194. 30 Oct. 2001.
[4] Lalithsena, Sarasi. "Domain-specific knowledge extraction from the web of data." (2018).
[5] Saif, Hassan et al. "Semantic sentiment analysis of twitter." *International semantic web conference*. Springer, Berlin, Heidelberg, 2012.
[6] Kumar, Abhishek et al. "Knowledge-enriched two-layered attention network for sentiment analysis." *arXiv preprint arXiv:1805.07819* (2018).
[7] Wang, Hongwei, et al. "DKN: Deep knowledge-aware network for news recommendation." *Proceedings of the 2018 world wide web conference*. 2018.
[8] Yang, Bishan, and Tom Mitchell. "Leveraging knowledge bases in lstms for improving machine reading." *arXiv preprint arXiv:1902.09091* (2019).
[9] Zhang, Fuzheng, et al. "Collaborative knowledge base embedding for recommender systems." *Proceedings of the 22nd ACM SIGKDD international conference on knowledge discovery and data mining*. 2016.

[10] Yu, Xiao, et al. "Personalized entity recommendation: A heterogeneous information network approach." *Proceedings of the 7th ACM international conference on Web search and data mining*. 2014.

[11] Zhao, Huan, et al. "Meta-graph based recommendation fusion over heterogeneous information networks." *Proceedings of the 23rd ACM SIGKDD International Conference on Knowledge Discovery and Data Mining*. 2017.

[12] Wang, Hongwei, et al. "Multi-task feature learning for knowledge graph enhanced recommendation." *The World Wide Web Conference*. 2019.

[13] Bhatt, Shreyansh, et al. "Knowledge graph enhanced community detection and characterization." *Proceedings of the Twelfth ACM International Conference on Web Search and Data Mining*. 2019.

[14] Bhatt, Shreyansh, et al. "Who Should Be the Captain This Week? Leveraging Inferred Diversity-Enhanced Crowd Wisdom for a Fantasy Premier League Captain Prediction." *Proceedings of the International AAAI Conference on Web and Social Media*. Vol. 13. No. 01. 2019.

[15] Nalamothu, Abhishek. "Abusive and Hate Speech Tweets Detection with Text Generation." (2019).

[16] Fang, Yuan, et al. "Object detection meets knowledge graphs." (2017).

[17] Sheth, Amit, et al. "Knowledge will propel machine understanding of content: extrapolating from current examples." *Proceedings of the International Conference on Web Intelligence*. 2017.

[18] Sheth, Amit at al. "Shades of Knowledge-Infused Learning for Enhancing Deep Learning." *IEEE Internet Computing* 23.6 (2019): 54-63.

Shreyash Bhatt is a Machine Learning Scientist at Amazon.com. Contact Shreyansh at bhattshr@amazon.com .

Amit Sheth is the director of Artificial Intelligence Institute at the University of South Carolina ( http://ai.sc.edu), and a fellow of IEEE, AAAI and AAAS. He can be reached at amit@sc.edu .

Valerie Shalin is a cognitive scientist and a professor of Psychology at Wright State University. She can be reached at valerie.shalin@wright.edu .

Jinjin Zhao is a Machine Learning Scientist at Amazon.com. Contact Jinjin at jinjzhao@amazon.com.